%% file: main.tex
\documentclass[10pt,twocolumn,letterpaper]{article}

\usepackage{cvpr}              
\usepackage{url}

\usepackage{latexsym}
\usepackage{microtype}

\usepackage{booktabs}
\usepackage{amsmath}
\usepackage{amssymb}
\usepackage{amsthm}

\newtheorem{theorem}{Theorem}

\usepackage{algorithm}
\usepackage{algpseudocode}
\usepackage{multirow}
\usepackage{mathtools}
\usepackage{subcaption} 
\usepackage{graphicx}
\usepackage{balance}
\usepackage{color}
\usepackage{xcolor}
\usepackage{enumitem}
\usepackage{xspace}

\newcommand{\stitle}[1]{\vspace{1mm} \noindent {\bf #1}}

\newcommand{\model}{StaR-KVQA}

\definecolor{mygreen}{rgb}{0.1, 0.6, 0.1}
\definecolor{myred}{rgb}{0.8, 0.1, 0.1}

\usepackage{caption}

\usepackage{type1cm} 
\usepackage{lettrine}

\usepackage{paracol} 
\usepackage{wrapfig}
\usepackage{graphicx}
\usepackage[table]{xcolor}  

\usepackage{amsthm}
\newtheorem{lemma}{Lemma}
\newtheorem{proposition}[theorem]{Proposition}

\definecolor{tableheaderbg}{gray}{0.92} 
\usepackage{ragged2e}
\usepackage{tcolorbox}
\tcbset{colback=white, colframe=black!15, boxsep=4pt, arc=1.5mm}

\tcbuselibrary{skins,breakable}

\definecolor{lightgray}{rgb}{0.95, 0.95, 0.95}
\definecolor{tableheaderbg}{RGB}{240,240,240}

\usepackage{array}
\input{preamble}
\definecolor{cvprblue}{rgb}{0.21,0.49,0.74}
\usepackage[pagebackref,breaklinks,colorlinks,allcolors=cvprblue]{hyperref}

\def\confName{CVPR}
\def\confYear{2026}

\title{StaR-KVQA: Structured Reasoning Traces for Implicit-Knowledge \\ Visual Question Answering}

\author{
Zhihao Wen$^{1}$\thanks{These authors contributed equally.},
Wenkang Wei$^{2}$\footnotemark[1],
Yuan Fang$^{3}$,
Xingtong Yu$^{3}$,
Hui Zhang$^{2,4}$\thanks{Corresponding authors.},
Weicheng Zhu$^{1}$,
Xin Zhang$^{1}$\footnotemark[2] \\
$^{1}$Ant International, Ant Group \\ 
$^{2}$School of Computer Science and Technology, University of Science and Technology of China \\ 
$^{3}$School of Computing and Information Systems, Singapore Management University  \\
$^{4}$Anhui Provincial Key Laboratory of High Performance Computing \\
{\tt\small \{z.wen, weicheng.zhu, evan.zx\}@antgroup.com} \\
{\tt\small yizhilouyi@mail.ustc.edu.cn, \{yfang, xingtongyu\}@smu.edu.sg, fzhh@ustc.edu.cn}
}

\begin{document}
\maketitle
\input{sec/sec-abstract}

\input{sec/sec-intro}
\input{sec/sec-related_work}
\input{sec/sec-prelimi}

\input{sec/sec-method}
\input{sec/sec-experiment}

\section{Conclusion}
We presented \textbf{\model}, which supervises dual symbolic relation paths and path-grounded explanations as \emph{structured reasoning traces}, turning reasoning from an implicit by-product into explicit, structured, and more transparent intermediate steps for IK-KVQA. Implemented as an implementation-friendly \emph{single-model} pipeline (\emph{dual-path planner} → \emph{reasoning composer} → \emph{best-triplet selector}, all with the same open-source MLLM), it requires no external retrievers/verifiers or extra trainable modules; traces are built offline and inference is a single autoregressive pass. With structure-aware self-distillation on trace-enriched data, \model\ attains strong answer accuracy and improved transparency of intermediate reasoning across benchmarks, achieving up to \textbf{+11.3\%} on OK-VQA over the strongest baseline and surpassing advanced closed-source systems (e.g., Gemini~2.5~Pro) under the IK-KVQA setting.
Remaining limitations include residual hallucination inherited from the backbone and the fact that, since the best-triplet selector is primarily optimized for answer-oriented consistency, the selected traces are not guaranteed to be the most intuitive or fully faithful explanations for humans. Faithfulness between explanations and paths is enforced only through coverage-based filtering and an LLM-as-a-judge selector, rather than formal guarantees, so occasional mismatches remain. Future work includes plug-compatible verification (e.g., retrieval-based checks or lightweight consistency modules), explicit objectives for explanation faithfulness and human preference, and broader cross-domain evaluations. Overall, \model\ advances more transparent multimodal reasoning for IK-KVQA while maintaining a simple, deployable pipeline.

\clearpage
\newpage

\section*{Acknowledgement}
The authors wish to thank Dr. Junnan Dong from Tencent Youtu Lab and Dr. Qinggang Zhang  from Hong Kong Polytechnic University for their valuable support of this work.

{
    \small
    \bibliographystyle{ieeenat_fullname}
    \bibliography{main}
}

\clearpage
\newpage

\appendix
\section*{Appendices}

\input{sec/sec-appendix}

\end{document}

%% file: sec/sec-abstract.tex
\begin{abstract}
Knowledge-based Visual Question Answering (KVQA) requires models to ground entities in images and reason over factual knowledge. Recent work has introduced its implicit-knowledge variant, \emph{IK-KVQA}, where a multimodal large language model (MLLM) is the sole knowledge source and answers are produced without external retrieval. 
Existing IK-KVQA approaches, however, are typically trained with answer-only supervision: reasoning remains implicit, justifications are often weak or inconsistent, and generalization after standard supervised fine-tuning (SFT) can be brittle.
We propose \textbf{\model}, a framework that equips IK-KVQA with \emph{dual-path structured reasoning traces}—symbolic relation paths over text and vision together with path-grounded natural-language explanations—to provide a stronger inductive bias than generic answer-only supervision. These traces act as modality-aware scaffolds that guide the model toward relevant entities and attributes, offering more structure than generic chain-of-thought supervision while not constraining reasoning to any single fixed path. With a single open-source MLLM, \model\ constructs and selects traces to build an offline trace-enriched dataset and then performs structure-aware self-distillation; no external retrievers, verifiers, or curated knowledge bases are used, and inference is a single autoregressive pass. Across benchmarks, \model\ consistently improves both answer accuracy and the transparency of intermediate reasoning, achieving up to \textbf{+11.3\%} higher answer accuracy on OK-VQA over the strongest baseline.
\end{abstract}

%% file: sec/sec-intro.tex
\section{Introduction}
\label{sec:intro}

Knowledge-based Visual Question Answering (KVQA) targets real-world scenarios where users ask questions about images that require factual knowledge beyond what is explicitly visible, and thus sits at the intersection of computer vision, natural language processing, and knowledge reasoning \citep{Wang2016FVQAFV, Marino2019OKVQAAV, Schwenk2022AOKVQAAB}.
Unlike conventional VQA that often learns a direct mapping from image features to textual answers, KVQA additionally requires \emph{grounding entities in the image} and \emph{linking them to relevant knowledge}.
For example, answering \emph{``Which breed of dog is this?''} involves recognizing visual cues (e.g., color, size) and associating them with prior knowledge about dog breeds.
In practical deployments, however, KVQA systems are often constrained by privacy/compliance, latency/cost, and reliability requirements, which can limit heavy external-retrieval pipelines and motivate more self-contained solutions.
The challenge is therefore not only perceiving pixels and text, but also \emph{organizing and using knowledge} in a way that improves answer quality under these constraints.

\begin{figure}[t]
   \centering
   \includegraphics[width=1\linewidth]{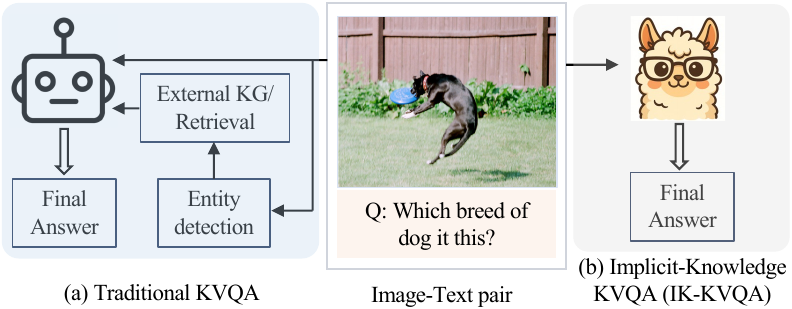}
   \caption{\textbf{Traditional KVQA vs.\ implicit-knowledge KVQA (IK-KVQA).}
Traditional KVQA often relies on external knowledge sources (e.g., retrieval or KGs) on top of a perception backbone.
In contrast, IK-KVQA retains the ``K'' to emphasize its knowledge-based nature while removing external sources: answers are predicted solely from $(I, Q)$ and parametric knowledge $f_\theta(I, Q)$.
}
   \label{fig:setting}
\end{figure}

Early KVQA systems often rely on explicit knowledge graphs (KGs) or retrieval modules \citep{Chen2024KnowledgeGM}.
While effective, such pipelines are not always a good fit for \emph{high-throughput} deployment.
First, external retrieval can introduce \textbf{privacy/compliance} risks when user images, queries, or extracted entities must be sent to third-party services or stored in external indices.
Second, retrieval and evidence fusion incur non-trivial \textbf{latency/cost} at scale, and performance can fluctuate with index freshness, domain shift, or infrastructure constraints.
Third, multi-stage designs reduce \textbf{reliability} and debuggability: errors in recognition or retrieval propagate, and evidence fusion can be brittle, making failures harder to attribute and audit.
These constraints motivate \emph{implicit-knowledge KVQA (IK-KVQA)} \citep{yang2022empirical}, where the task remains knowledge-based but external sources are disallowed: multimodal large language models
(MLLMs)\footnote{In the literature, models such as Qwen2.5-VL are also referred to as VLMs. We use the term \emph{MLLMs} in this paper for consistency.}
must answer directly from $(I,Q)$ by leveraging \emph{parametric} knowledge, as illustrated in Figure~\ref{fig:setting}.
Importantly, IK-KVQA is not meant to replace KB/RAG-based KVQA; rather, it captures a practically common regime where the system must be \emph{self-contained}, \emph{cost-effective}, and \emph{auditable}.
In this stricter setting, the bottleneck shifts from retrieving knowledge to \emph{eliciting, structuring, and validating} the model's internal knowledge so that it supports accurate predictions without relying on opaque shortcuts.

\stitle{Challenges and Our Approach.}
The IK-KVQA setting simplifies system design and removes external dependencies, but also imposes stricter demands: the model must rely solely on its parameters to ground evidence, recall factual knowledge, and reason.
In practice, MLLMs often behave as \emph{black boxes}---sometimes producing correct answers while intermediate descriptions are \emph{underspecified, weakly grounded, or inconsistent}.
The absence of explicit, structured supervision on intermediate steps complicates analysis and can affect reliability.
Concretely, IK-KVQA faces three core challenges:
(1) \emph{Lack of explicit supervision}, since models are typically trained only on final answers while reasoning traces remain hidden;
(2) \emph{Underspecified intermediate signals}, where predictions may lack consistently aligned stepwise descriptions; and 
(3) \emph{Potential overfitting}, as conventional fine-tuning can bias toward in-domain patterns with reduced robustness beyond the training distribution.

To address these issues, we propose \textbf{\model}, which equips MLLMs with \emph{dual-path structured reasoning traces}.
Instead of leaving reasoning implicit, \model\ supervises both symbolic relation paths and natural-language explanations, so training better reflects how models should connect visual cues with internal knowledge.
We use \emph{relation paths} as planning scaffolds: relations are more stable than surface entities, share a compact ontology across text and vision, and align naturally with object- and scene-level attributes.
These traces serve as \emph{soft plans} that highlight salient entities/attributes while keeping generation flexible.

\model\ reuses a single open-source MLLM (e.g., Qwen2.5-VL-7B) to generate dual relation paths, compose explanations, and select the most consistent triplet, yielding an augmented dataset with explicit traces.
Fine-tuning on it performs \emph{structure-aware self-distillation}, learning from both final answers and intermediate signals (paths + explanations).
This supervision provides a stronger inductive bias, reducing shortcut reliance and improving accuracy.
At inference, the fine-tuned model generates traces and answers in a single autoregressive pass without external knowledge.
Overall, \model\ extends self-distillation to multimodal reasoning by distilling \emph{structured intermediate reasoning} rather than answer-only outputs.
In summary, our contributions are threefold:
\begin{itemize}
    \item \textbf{Structured supervision for IK-KVQA.}
    We introduce \textbf{\model}, replacing answer-only supervision with \emph{structured reasoning traces}: dual symbolic relation paths and path-grounded explanations as modality-aware scaffolds, without constraining reasoning to a single path.
    
    \item \textbf{Single-model, dependency-free pipeline.}
    We develop an implementation-friendly pipeline---\emph{dual-path planner}, \emph{reasoning composer}, and an \emph{internal selector} instantiated by the same model---to construct trace-enriched data for \emph{structure-aware self-distillation}.
    The system uses a \textbf{single} open-source MLLM, adds no retrievers/verifiers or extra trainable modules, and keeps inference to one pass.
    
    \item \textbf{Empirical gains in accuracy and transparency.}
    Fine-tuning on trace-enriched data consistently improves accuracy and intermediate-trace transparency across benchmarks (e.g., up to \textbf{+11.3\%} on OK-VQA over the strongest baseline), while remaining fully in the no-retrieval setting.
\end{itemize}

%% file: sec/sec-related_work.tex
\section{Related Work}
\label{sec:related_work}

We review KVQA with retrieval, KVQA with LLMs \ / MLLMs, and self-distillation, and position our contributions.

\textbf{KVQA with knowledge graphs or retrieval.}
Early datasets (FVQA \citep{Wang2016FVQAFV}, OK-VQA \citep{Marino2019OKVQAAV, schwenk2022okvqa}, KVQA \citep{shah2019kvqa}) spurred pipelines that integrate explicit KGs or retrievers, e.g., ConceptBERT \citep{garderes2020conceptbert}, MAVEx \citep{wu2022multi}, and KRISP \citep{marino2021krisp}.
More recent retrieval-augmented systems such as Wiki-LLaVA \citep{caffagni2024wiki}, RoRA-VLM \citep{qi2024rora}, and EchoSight \citep{yan2024echosight} further demonstrate the value of external knowledge, but introduce pipeline complexity, error propagation, and maintenance costs with limited transparency.
Most still rely on answer supervision, fusing retrieved facts into hidden representations rather than supervising explicit, auditable reasoning traces.

\textbf{KVQA with LLMs \ / MLLMs.}
To reduce reliance on explicit KGs, LLMs have been used as implicit knowledge engines: PICa \citep{yang2022empirical} shows GPT-3 \citep{brown2020language} can answer knowledge-intensive questions from captions; KAT \citep{Gui2021KATAK} and REVIVE \citep{Lin2022REVIVERV} add supporting evidence, while MAIL \citep{dong-etal-2024-modality} and ReflectiVA \citep{cocchi2025augmenting} explore reflective or adaptive fusion.
However, reasoning traces in these systems are often absent or weakly grounded, limiting interpretability and fine-grained control of knowledge use.
Recent MLLMs perform end-to-end image--text reasoning via lightweight projections \citep{liu2023visual, liu2024improved}, Q-Former-style modules \citep{li2023blip, dai2023instructblip}, Perceiver-style encoders \citep{laurenccon2024matters}, or cross-attention as in Flamingo \citep{alayrac2022flamingo, awadalla2023openflamingo}.
Training typically combines large-scale caption alignment \citep{changpinyo2021conceptual, gadre2023datacomp, laurenccon2024matters} with visual instruction tuning \citep{laurenccon2023obelics}.
In the IK-KVQA regime, their reasoning remains largely implicit and weakly supervised.

\textbf{Self-distillation and reasoning supervision.}
Self-distillation \citep{Zhang2019BeYO, Zhang2021SelfDistillationTE} uses model outputs as auxiliary supervision; SDFT \citep{Yang2024SelfDistillationBD} rewrites responses to mitigate forgetting, and \citet{wang2023unifying} add structural signals for multi-hop QA.
Yet most works distill only answers, leaving reasoning implicit.
In contrast, we distill \emph{structured reasoning traces}---dual symbolic paths and path-grounded explanations---as explicit supervision for IK-KVQA in a single-model, parametric-only setting, enabling self-distillation with more transparent intermediate reasoning.

\begin{figure*}[t]
   \centering
   \includegraphics[width=0.9\linewidth]{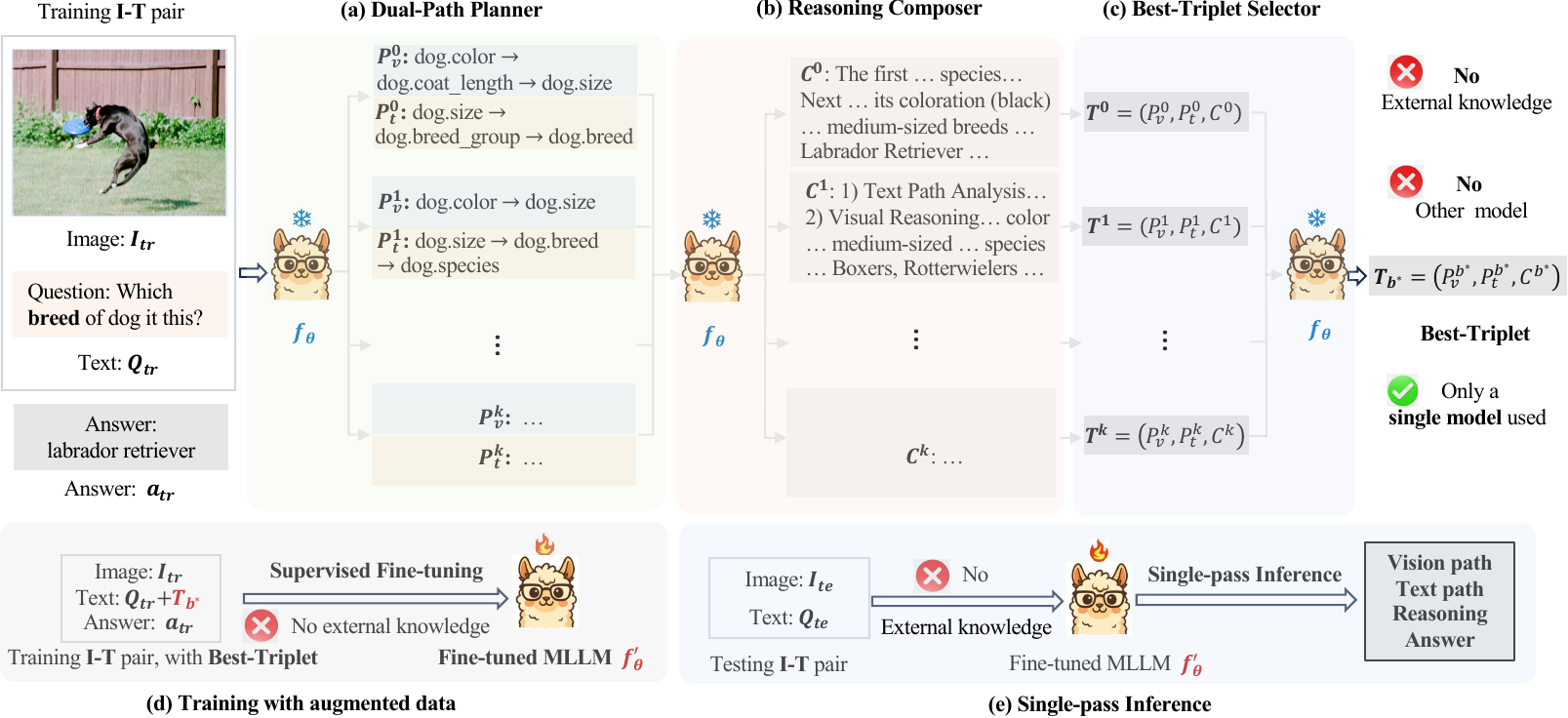}
\caption{
\textbf{Overview of \model.} 
Given a training image–text pair, a single $\mathrm{MLLM}_{\phi}$ generates multiple dual relation paths (a) and corresponding explanations (b). A selector (c) identifies the most consistent triplet, which, combined with the ground-truth answer, forms reasoning-augmented supervision (d). The fine-tuned $f_\theta'$ then performs single-pass inference (e), producing reasoning traces and answers without external knowledge. 
An example dual-path scaffold is shown, highlighting relevant visual attributes and semantic priors; paths need not be minimal or sufficient but guide the model toward useful evidence before composing a full explanation.
}
\vspace{-4mm}
\label{fig:framework}
\end{figure*}

%% file: sec/sec-prelimi.tex
\section{Preliminaries}
\label{sec:prelim}

We formalize the IK-KVQA setting, and task notation.

\stitle{Problem definition.}
Given an image $I$ and a question $Q$, knowledge-based visual question answering (KVQA) aims to predict an answer $\hat a \in \mathcal{A}$:
\begin{align}
\hat a = f(I, Q, K),
\end{align}
where $f$ is the answering model and $K$ denotes external knowledge retrieved from a knowledge graph or textual corpus.
Traditional KVQA pipelines typically ground entities in the image and query $K$ to supplement factual reasoning.

\stitle{Implicit-knowledge KVQA.}
In contrast, we consider the implicit-knowledge setting (IK-KVQA), where $K$ is unavailable.
The only information sources are (i) visual evidence from $I$, (ii) linguistic cues from $Q$, and (iii) \emph{parametric knowledge} encoded in model parameters.
Under this setting, the answer is predicted as
\begin{align}
\hat a = f_{\theta}(I, Q),
\end{align}
where $f_\theta$ is trained solely without external retrieval.
This formulation removes external dependencies but leaves intermediate reasoning largely implicit.
Our framework addresses this gap by augmenting supervision with explicit reasoning traces (dual paths and natural-language explanations), keeping all computation within the single model.

%% file: sec/sec-method.tex
\section{Structured Reasoning Traces for IK-KVQA}
\label{sec:method}

We introduce \textbf{\model}, which replaces answer-only supervision with \emph{structured reasoning traces}: dual relation paths $(P_t, P_v)$ and a path-grounded explanation $C$. This converts reasoning from an implicit by-product into explicit, structured training signals. The entire pipeline runs within a single $\mathrm{MLLM}_{\phi}$, which generates paths, composes explanations, and selects the best triplet, all without external retrievers, verifiers, or curated knowledge bases (Figure~\ref{fig:framework}).

\noindent\textbf{Design principles.}
(i) \emph{Inductive structure:} relation paths act as low-dimensional planning scaffolds for cross-modal reasoning; 
(ii) \emph{Trace-level consistency:} the paths$\rightarrow$explanation$\rightarrow$answer pipeline encourages alignment between intermediate traces and final predictions; 
(iii) \emph{Single-family learning:} generation and supervision remain style-aligned via self-distillation within the same MLLM family.

\subsection{Dual-Path Planner}
To explicitly structure reasoning across linguistic and visual modalities, we design a \textbf{dual-path planner} that generates symbolic \emph{relation paths}.  
These paths capture semantic relations between entities and attributes, and are commonly used in knowledge-graph reasoning due to their stability and interpretability \citep{wang2021relational, xu2022subgraph, wang2023survey, wen2023augmenting, wen2024prompt}.  
Unlike dynamic entities, \textbf{relations are more stable and reusable}, making them low-dimensional, discrete surrogates for reasoning \textbf{scaffolds} rather than exact proofs.

Formally, given an image–question pair $(I, Q)$, the frozen backbone $\mathrm{MLLM}_{\phi}$ generates $K$ candidate path pairs:
\begin{align}
    \{(P_t^{(k)}, P_v^{(k)})\}_{k=1}^K = \mathrm{Planner}_{\phi}(I, Q),
\end{align}
where each $(P_t^{(k)}, P_v^{(k)})$ consists of: (1) a \emph{text path} $P_t^{(k)}$ capturing semantic associations from $Q$ and linguistic priors, and (2) a \emph{vision path} $P_v^{(k)}$ encoding attributes and relations grounded in $I$.  

These dual paths serve as \emph{soft planning hints}: they guide which entities and attributes to consider, without rigidly constraining the reasoning chain. And the downstream reasoning composer (Section~\ref{sec:method:reasoning_composer}) can incorporate additional cues or skip redundant links for final answers.

We operationalize \emph{plan-then-solve} ideas for IK-KVQA by planning \emph{internally} over relation paths within a \textbf{single-model} setup, avoiding external KGs or retrieval.  
Unlike prior plan-first prompting \citep{Wang2023PlanandSolvePI} or KG path reasoning \citep{Luo2023ReasoningOG}, our planner unifies textual priors and visual attributes into dual relation paths inside a single-model pipeline. This approach provides multiple candidate routes and acts as an \emph{inductive bias}, narrowing the search space while maintaining auditable intermediate steps.

For example, consider the question ``\texttt{Which breed of dog is this?}'' with image $I$.  
One candidate might be:  
``$P_v^{(k)}$: \texttt{dog.color → dog.coat\_length → dog.size},  
$P_t^{(k)}$: \texttt{dog.size → dog.breed\_group → dog.breed},'' as shown in Figure~\ref{fig:framework} (a).  
These complementary relation paths connect visual cues with semantic priors, steering the model away from label-memorization shortcuts.  
In practice, paths can include redundant or mildly spurious hops; we treat them as noisy but useful scaffolds. The best-triplet selector and self-distillation (Sections~\ref{sec:method:best_triplet_selector}–\ref{sec:method:training_aug}) further refine these paths, preferring those that consistently support correct answers.

\subsection{Reasoning Composer}
\label{sec:method:reasoning_composer}
Given a dual-path pair $(P_t^k, P_v^k)$, the \textbf{reasoning composer} turns abstract plans into natural-language \emph{reasoning content} $C^k$ using the \textbf{same} backbone:
\begin{align}
    C^k = \mathrm{Compose}_{\phi}(I, Q, P_t^k, P_v^k).
\end{align}
We build on evidence that explanations can act as supervision: VQA-NLE-style rationales improve answer quality and interpretability \citep{Suo2023S3CSV,Irawan2024TowardsEA,Xie2024KnowledgeAugmentedVQ}; chain-of-thought prompts in ScienceQA induce more structured reasoning \citep{Lu2022LearnTE,Zhang2023MultimodalCR}; and explicit clues or explanation–answer agreement (DCLUB, MCLE) reduce shortcutting and inconsistency \citep{fu2023dynamic,Lai2023TowardsMF}.  

Our composer instantiates these insights \emph{within the IK setting} by explicitly \emph{binding} the rationale to the proposed paths. During trace construction, we instruct $\mathrm{Compose}_{\phi}$ to (i) mention at least one attribute or relation token from $P_v^k$ and (ii) include at least one semantic hop from $P_t^k$ in $C^k$. We then compute a simple coverage score between the tokenized explanation and the path tokens, and discard candidates with very low coverage (e.g., no overlap on either path). The best-triplet selector (Section~\ref{sec:method:best_triplet_selector}) is applied after this filtering and further down-weights explanations that only weakly cite elements from $P_t^k$/$P_v^k$. This conditioning discourages free-form but ungrounded narratives and keeps explanations focused on the entities and attributes used in the symbolic plan.  
This binding makes explanations easier to audit against the paths while still allowing additional cues. In practice, it turns explanation quality into a path-aware supervision signal that the \textbf{single-model} system can learn from, rather than treating interpretability as a purely post-hoc by-product.

\subsection{Best-Triplet Selector}
\label{sec:method:best_triplet_selector}
Not all triplets $(P_t^k, P_v^k, C^k)$ are reliable, and directly using them may introduce noisy or inconsistent supervision.  
We therefore introduce a \textbf{best-triplet selector} that filters candidates during the \emph{data augmentation stage}, where dual paths and reasoning contents are turned into training signals.

The selector is instantiated as an \emph{LLM-as-a-judge} within the same \textbf{single-model} setup, reusing $\mathrm{MLLM}_{\phi}$.  
Given $(I, Q)$ and a set of candidates $\{(P_t^k, P_v^k, C^k)\}_{k=1}^K$, we prompt $\mathrm{MLLM}_{\phi}$ to rank triplets according to three criteria:  
(i) \emph{answer-oriented path consistency} (the answer naturally follows from the explanation and paths),  
(ii) \emph{internal coherence and conciseness}, and  
(iii) \emph{path citation} (explicitly mentioning elements from $P_t^k$/$P_v^k$).  
The primary objective is answer quality: the selector prefers triplets for which the predicted answer is well supported by the textual explanation and dual paths, while faithfulness is encouraged but not enforced as a hard constraint.  
Formally, with score $s_{\phi}$:
\begin{align}
b^{*} &= \underset{b}{\arg\max}\; s_{\phi}(I, Q, P_t^b, P_v^b, C^b), \nonumber\\
T_{b^{*}} &= (P_t^{b^{*}}, P_v^{b^{*}}, C^{b^{*}}).
\end{align}
This step adds \emph{no additional trainable parameters}.  
Reusing the same backbone keeps $\mathcal{D}_{\text{aug}}$ style-aligned with the model’s own traces.  
The selected triplet reflects what the MLLM \emph{internally} finds most helpful for answering the question—possibly not the most intuitive chain for humans, but empirically providing stronger supervision in a lightweight, fully parametric pipeline.

\stitle{Why single-model trace construction (vs.\ extra modules)?}
We avoid extra verifiers or retrievers because:
(i) \emph{Homogeneous generation–learning:} planning, composing, and selecting all use the same model family, so the student $f_\theta$ learns from in-family traces, mitigating supervision–generation mismatch and catastrophic forgetting \citep{Yang2024SelfDistillationBD};
(ii) \emph{Test-time simplicity:} selection is used only offline during augmentation, leaving inference as a single autoregressive pass;
(iii) \emph{IK compliance:} the design stays fully parametric, with no external knowledge or modules.

\subsection{Training with Augmented Data}
\label{sec:method:training_aug}
Let the training split be $\{(I_{tr}^i, Q_{tr}^i, a_{tr}^i)\}_{i=1}^N$, 
where $I_{tr}^i$ is the image, $Q_{tr}^i$ the question, and $a_{tr}^i$ the ground-truth answer.  
For each pair $(I_{tr}^i, Q_{tr}^i)$, the planner and composer generate multiple candidate triplets, and the selector chooses the best one $T^i_{b^*} = (P_t^{b^*}, P_v^{b^*}, C^{b^*})$.  
We then construct the \textbf{augmented training set}:
\begin{align}
\mathcal{D}_{\text{aug}} = \{(I_{tr}^i, Q_{tr}^i, T^i_{b^{*}}, a_{tr}^i)\}_{i=1}^N.
\end{align}

The base model $f_\theta$ is fine-tuned on $\mathcal{D}_{\text{aug}}$ with a token-level cross-entropy loss: 
\begin{align}
\mathcal{L}_{\text{SFT}}(\theta;\mathcal{D}_{\text{aug}}) 
= - \sum_{(I, Q, T, a)\in \mathcal{D}_{\text{aug}}} 
\log p_\theta(T, a \mid I, Q),
\end{align}
where the target sequence concatenates the reasoning paths $P_t, P_v$, reasoning content $C$, and the final answer $a$.  
This objective encourages the fine-tuned model $f_\theta'$ to jointly generate structured reasoning traces and correct answers in a single coherent output.

\subsection{Single-pass Inference}
At test time, given $(I_{te}, Q_{te})$, the fine-tuned model $f_\theta'$ performs a \emph{single} autoregressive decode that jointly emits dual paths, a path-grounded explanation, and the final answer:
\begin{align}
f_\theta'(I_{te}, Q_{te}) = (\hat{P}_t, \hat{P}_v, \hat{C}, \hat{a}).
\end{align}
No selector or auxiliary module is invoked at inference, and the paths$\rightarrow$explanation$\rightarrow$answer structure directly exposes a complete trace for auditing, without any external retrieval.

\begin{table*}[h!]
\scriptsize
\renewcommand{\arraystretch}{0.95} 
\setlength{\tabcolsep}{16pt} 
\centering
\caption{
\centering
Performance comparison on OK-VQA.  
}
\label{tab:exp main okvqa}
\begin{tabular}{c|c|c|c}
\hline
\textbf{Method}                & \textbf{Model Inputs}            & \textbf{External Knowledge}          & \textbf{Acc. (\%)} \\ \hline
Q Only                        & Question + Image                 & -                                                     & 14.93              \\ \hline
\multicolumn{4}{c}{\textbf{KVQA with Knowledge Graphs and Retrieval}}                                                 \\ \hline
BAN                            & Question + Image                 & -                                                          & 25.17  \\        
BAN +AN                        & Question + Image                 & Wikipedia                                       & 25.61              \\ 
MUTAN                          & Question + Image                 & -                                                    & 26.41              \\ 
MUTAN +AN                      & Question + Image                 & Wikipedia                                       & 27.84              \\ 
ConceptBERT                    & Question + Image                 & ConceptNet                                      & 33.66              \\ 
HCNMN                          & Question + Image                 & WordNet                                         & 36.74              \\ 
Krisp                          & Question + Image                 & Wikipedia + ConceptNet                          & 38.90              \\ 
MAVEx                          & Question + Image                 & Wikipedia + ConceptNet + Google Images         & 41.37              \\ 
VLC-BERT                       & Question + Image                 & COMET + ConceptNet                              & 43.14              \\ 
MCAN                           & Question + Image                                             & -                             & 44.65              \\ 
\hline
\multicolumn{4}{c}{\textbf{KVQA with LLMs \ / MLLMs}}                                          \\ \hline
PICA-Base                      & Question + Caption + Object Tags & Frozen GPT-3 (175B)                                           & 43.30              \\ 
Pica-Full                      & Question + Caption + Object Tags & Frozen GPT-3 (175B)                                            & 48.00              \\ 
KAT (Single)                   & Question + Caption + Object Tags & Frozen GPT-3 (175B) + Wikidata                  & 53.09              \\ 
KAT (Ensemble)                 & Question + Caption + Object Tags & Frozen GPT-3 (175B) + Wikidata                  & 54.41              \\ 
REVIVE                         & Question + Caption + Region Tags & Frozen GPT-3 (175B) + Wikidata                  & 53.83              \\ 
\hline
MAIL                   & Question + Image                 & Frozen MiniGPT-4 (7B) + ConceptNet    & 56.69   \\ 
\hline

\multicolumn{4}{c}{\textbf{IK-KVQA with MLLMs}}                                          \\ \hline
Qwen2.5-VL-7B                      & Question + Image & Qwen2.5-VL-7B                              &    75.74          \\ 
Llama-3.2-11B-Vision               & Question + Image & Llama-3.2-11B-Vision                       &    67.84          \\ 
Gemma-3-12B                      & Question + Image & Gemma-3-12B                                  &    71.40         \\ 
Gemma-3-27B                    & Question + Image & Gemma-3-27B                             &    79.34          \\ 
Qwen2.5-VL-72B                      & Question + Image & Qwen2.5-VL-72B                           &  80.75            \\ 
InternVL3-78B                   & Question + Image &InternVL3-78B                              &   67.61           \\ 
\hline
GPT-4o                  & Question + Image & GPT-4o                                                &  77.86            \\ 
Gemini 2.5 Flash   & Question + Image &Gemini 2.5 Flash                                            &  79.97           \\ 
Gemini 2.5 Pro     & Question + Image &Gemini 2.5 Pro                                              &  80.53           \\ 
\hline

SFT & Question + Image & Fine-tuned Qwen2.5-VL-7B         & 76.36              \\
CoT                   & Question + Image &  Qwen2.5-VL-7B                   & 76.88             \\ 
CoT + SFT & Question + Image & Fine-tuned Qwen2.5-VL-7B & 79.58 \\
LLaVA-CoT & Question + Image & Fine-tuned Llama-3.2-11B-Vision  &  76.57            \\ 
M2-Reasoning             & Question + Image & M2-Reasoning-7B   & 78.63         \\ 
SDFT & Question + Image &Fine-tuned Qwen2.5-VL-7B                   &\underline{82.56} \\
\hline
\model$_{Qwen}$   & Question + Image &Fine-tuned Qwen2.5-VL-7B           & \textbf{91.51}             \\ 
\model$_{Llama}$  & Question + Image &Fine-tuned Llama-3.2-11B-Vision    & \textbf{90.01}             \\ 
\model$_{Gemma}$  & Question + Image &Fine-tuned Gemma-3-12B             & \textbf{91.90}           \\ 
\hline
\end{tabular}
\end{table*}

%% file: sec/sec-experiment.tex
\section{Experiments}
\label{sec:expriments}

We conduct extensive experiments to evaluate \textbf{\model}, with comparison to state-of-the-art baselines and in-depth model analysis.

\subsection{Experimental Setup}

\stitle{Datasets.}
In line with recent advances in the field \citep{Marino2019OKVQAAV, Yang2021AnES, Gui2021KATAK, wu2022multi, Lin2022REVIVERV}, we performed our primary validation on the OK-VQA dataset. Comprising 14,055 image-question pairs, this benchmark is currently the most demanding in the domain. Furthermore, to establish the broader applicability of our model, we performed supplementary experiments on FVQA \citep{Wang2016FVQAFV}, which initiated the exploration of KVQA.

\stitle{Baselines.}
Three categories.
(i) \textbf{KVQA+KG/Retrieval}: Q Only~\citep{Marino2019OKVQAAV}, BAN~\citep{Kim2018BilinearAN}, MUTAN~\citep{Benyounes2017MUTANMT}, ConceptBERT~\citep{garderes2020conceptbert}, KRISP~\citep{marino2021krisp}, MAVEx~\citep{wu2022multi}, VLCBERT~\citep{Ravi2022VLCBERTVQ}, HCNMN~\citep{Zhang2023TowardMD}, MCAN~\citep{Yu2019DeepMC}; BAN/MUTAN use ArticleNet~\citep{Marino2019OKVQAAV}.
(ii) \textbf{KVQA+LLMs}: PICa~\citep{Yang2021AnES}, KAT~\citep{Gui2021KATAK}, REVIVE~\citep{Lin2022REVIVERV}.
(iii) \textbf{IK-KVQA+MLLMs}: open-source Qwen2.5-VL-7B/72B~\citep{Bai2025Qwen25VLTR}, Llama-3.2-11B-Vision~\citep{Dubey2024TheL3}, Gemma-3-12B/27B~\citep{Kamath2025Gemma3T}, InternVL3-78B~\citep{Zhu2025InternVL3EA}; proprietary Gemini 2.5 Flash/Pro~\citep{Comanici2025Gemini2P}, GPT-4o~\citep{Hurst2024GPT4oSC}; SFT~\citep{ouyang2022training}, CoT~\citep{wei2022chain}, LLaVA-CoT~\citep{xu2024llava}, M2-Reasoning (7B)~\citep{ai2025m2}, SDFT~\citep{Yang2024SelfDistillationBD}. SFT, CoT, CoT + SFT, and SDFT are augmented on Qwen2.5-VL-7B.
``\textit{CoT + SFT}" is a well-optimized CoT-prompted SFT baseline. 
All MLLMs are \textbf{instruction-tuned}. Results for (i)(ii) follow \citet{dong-etal-2024-modality}.

\stitle{Protocol:}
Seed~42; default decoding; no CoT is used unless specified ; inputs=(image, question) only for IK-KVQA approaches; single-run reporting per \citet{dong-etal-2024-modality}.
\stitle{Implementation:}
\model\ is trained in PyTorch~2.7.0, Python~3.10 on NVIDIA L20; batch (accum.) $16$, LR $1\mathrm{e}{-4}$, LoRA rank $32$, alpha $64$, $3$ epochs; $K{=}3$ (OK-VQA), $K{=}4$ (FVQA).The backbone MLLMs include \textbf{Qwen2.5-VL-7B}, \textbf{Llama-3.2-11B-Vision}, and \textbf{Gemma-3-12B}.
\stitle{Metric:}
Direct-answer VQA accuracy~\citep{Agrawal2015VQAVQ}:
$\text{Acc}=\min \left(\frac{\#\text{humans with that answer}}{3},1\right).
$
Normalize (lowercase, digits, remove punctuation/articles).

\begin{table}[t]
  \centering
  \scriptsize
  \renewcommand{\arraystretch}{0.95} 
  \addtolength{\tabcolsep}{16pt}
  \caption{Performance comparison of IK-KVQA with MLLMs approaches on \textbf{FVQA}. }
  \label{tab:exp main fvqa}
\centering
\begin{tabular}{c|c}
      \toprule
      \textbf{Method} &\textbf{Acc. (\%)}  \\
      \midrule
       Qwen2.5-VL-7B    & 71.61      \\ 
       Llama-3.2-11B-Vision   & 66.09  \\ 
       Gemma-3-12B            & 70.64 \\ 
       Gemma-3-27B  & 76.82 \\
       Qwen2.5-VL-72B & 75.95 \\
       InternVL3-78B & 70.99 \\
       \hline
       GPT-4o          &  72.36            \\ 
Gemini 2.5 Flash       &  74.51           \\ 
Gemini 2.5 Pro         &  73.39           \\ 
       \hline
       SFT & 73.91  \\
       CoT & 74.66 \\
       CoT + SFT & 75.13\\
       LLaVA-CoT & \underline{78.45}  \\
       M2-Reasoning & 72.53 \\
       SDFT & 75.54  \\
       \hline
       \model$_{Qwen}$   & \textbf{82.82}             \\ 
        \model$_{Llama}$  & \textbf{80.19}             \\ 
        \model$_{Gemma}$  & \textbf{81.20}           \\ 
      \bottomrule
    \end{tabular}
\end{table}

\begin{table*}[tbp]
    \centering
    \footnotesize
    \caption{Ablation studies.\\
    }
    \label{table:ablation}%
    \renewcommand{\arraystretch}{0.9} 
    \setlength{\tabcolsep}{8pt} 
    \begin{tabular}{@{}l|cccc|ccc|ccc|c@{}}
    \toprule
    \multirow{2}*{Variants}
    & Vision & Text & Reasoning & Best-Triplet  &\multicolumn{3}{c|}{OK-VQA} &\multicolumn{3}{c|}{FVQA} & \multirow{2}*{Average} \\
    &Path &Path &Composer & Selector & Qwen & Llama & Gemma & Qwen & Llama & Gemma  \\
    \midrule
    No paths
    & $\times$ & $\times$ & $\checkmark$ & $\checkmark$   &87.47&72.57&89.09  &76.31&\underline{76.31}&79.14 &80.15 \\
    No content
    & $\checkmark$ & $\checkmark$ & $\times$ & $\checkmark$  &87.53&\underline{86.00}&88.26  &76.22&76.05&73.34 &\underline{81.23}\\
    No text path
    & $\checkmark$ & $\times$ & $\checkmark$ & $\checkmark$  &83.66&72.77&87.84  &76.91&56.91&79.66 & 76.29\\  
    No vision path
    & $\times$ & $\checkmark$  & $\checkmark$ & $\checkmark$  &\textbf{92.65}&70.01&86.92  &74.42&64.81&78.20 & 77.84\\
    No selector
    & $\checkmark$ & $\checkmark$ & $\checkmark$ & $\times$   & \underline{91.76}&72.17&\textbf{91.94}  &\textbf{84.55}&49.18&\textbf{83.18} & 78.80\\
    \midrule
    \model
    & $\checkmark$ & $\checkmark$ & $\checkmark$ & $\checkmark$  &91.51&\textbf{90.01}&\underline{91.90}  &\underline{82.82}&\textbf{80.19}&\underline{81.20}  & \textbf{86.27}\\
    \bottomrule
    \end{tabular}
\end{table*}

\subsection{Main Results}
\label{sec:exp main_results}
We report comparisons with representative baselines in Table~\ref{tab:exp main okvqa} and Table~\ref{tab:exp main fvqa}. 
Several key observations emerge:  
(i) \textbf{MLLMs as strong backbones.}  
Methods based on state-of-the-art multimodal large language models (MLLMs) achieve the strongest overall performance, even without explicit external knowledge. 
This confirms that parametric knowledge acquired in large-scale pretraining is already highly effective for KVQA, while being simpler to use than retrieval- or KG-based approaches.  
(ii) \textbf{\model\ achieves the best results.}  
Among the MLLM-based methods, our reasoning-augmented framework consistently delivers the best performance.  
On OK-VQA, it surpasses the strongest baseline by up to \textbf{+11.3\%}, highlighting the effectiveness of augmenting training with \emph{structured reasoning traces}.  
(iii) \textbf{Closed-source models are strong but surpassed.}  
Closed-source commercial systems achieve competitive results but still fall short of our approach.  
Notably, \model\ outperforms \textbf{Gemini 2.5 Pro}, one of the most advanced multimodal reasoning models in our comparison.  
(iv) \textbf{Self-distillation is strong but limited.}  
We also evaluate \textbf{Self-Distillation Fine-Tuning (SDFT)} \citep{Yang2024SelfDistillationBD}, which rewrites task responses into the model’s own style for fine-tuning.  
With Qwen2.5-VL-7B as the backbone, SDFT already exceeds Gemini 2.5 Pro by over 2\% on OK-VQA and ranks just below our method, underscoring the strength of self-distillation in the IK-KVQA regime.  
\model\ goes further: by supervising both symbolic paths and natural-language explanations as \emph{structured reasoning traces}, it retains SDFT’s accuracy gains while providing more transparent intermediate reasoning.  

In summary, \model\ not only surpasses strong open-source and closed-source baselines but also sets a new state of the art in IK-KVQA, combining superior accuracy with more interpretable reasoning behavior.

\subsection{Ablation Studies \& Hyperparameters}

\stitle{Ablation studies.}
We conduct ablations to examine the role of each component in our reasoning-augmented framework (Table~\ref{table:ablation}). 
For each variant, we regenerate the augmented training data and retrain the model so that the reported performance reflects the absence of the removed component.  
Removing either the dual paths (\textbf{no paths}) or the explanations (\textbf{no content}) leads to clear accuracy drops, confirming that symbolic paths and natural-language reasoning provide complementary supervision. Restricting the framework to a single modality (\textbf{vision-only} or \textbf{text-only}) further degrades performance, underscoring the need to align textual priors with visual grounding.
Replacing the best-triplet selector with random selection (\textbf{no selector}) yields mixed outcomes: it can slightly improve Qwen and Gemma on some datasets but severely harms LLaMA, indicating that the selector is important for robustness across backbones even if random choice occasionally preserves strong candidates.  
Overall, these ablations show that dual paths, reasoning content, and the selector all contribute and, in combination, explain why the full \textbf{\model} model delivers strong and balanced performance across benchmarks.

\begin{figure}[t]
  \centering
  \begin{subfigure}[b]{0.48\linewidth}
    \centering
    \includegraphics[width=\linewidth]{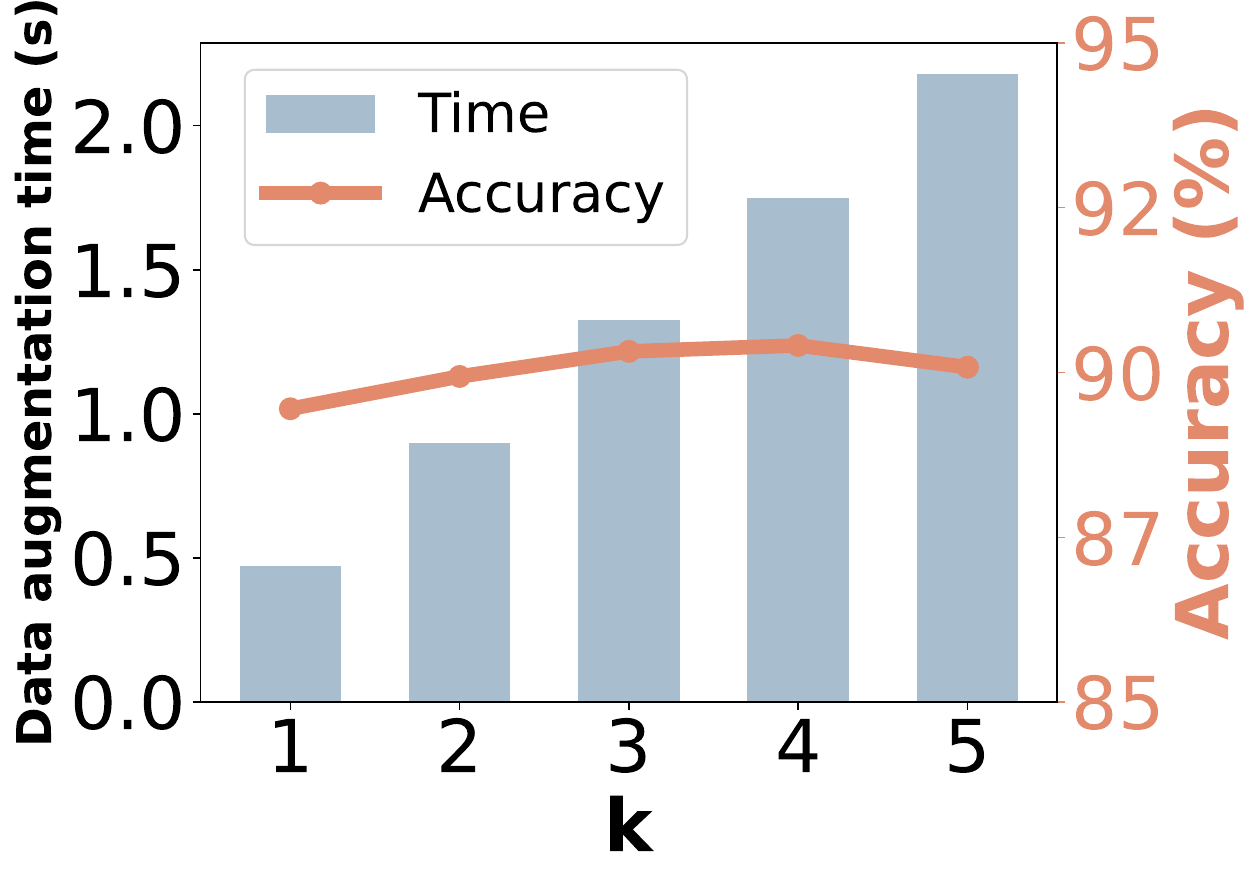}
    \caption{OK-VQA}
  \end{subfigure}
  \hfill
  \begin{subfigure}[b]{0.48\linewidth}
    \centering
    \includegraphics[width=\linewidth]{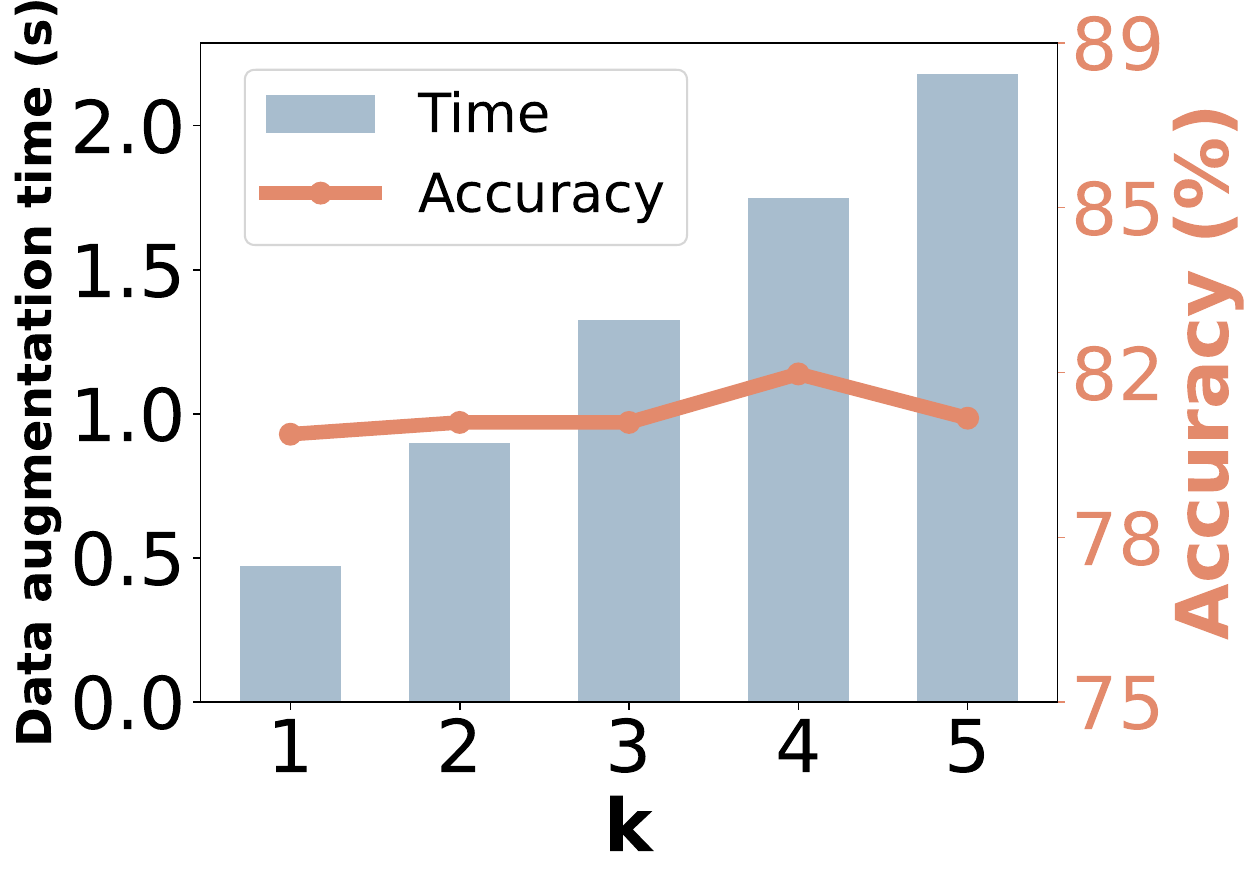}
    \caption{FVQA}
  \end{subfigure}
  \caption{$K$, the number of candidate paths.}
  \label{fig:hyperparameters}
\end{figure}

\stitle{Hyperparameters.}  
We next study the sensitivity to the number of candidate paths $K$, using Qwen2.5-VL-7B as the backbone. 
For each $K$, we report both answer accuracy and the average time cost of running the full \model\ augmentation pipeline per training example.  
As shown in Figure~\ref{fig:hyperparameters}, increasing $K$ initially improves performance by providing richer reasoning options, but when $K$ becomes too large (e.g., $K=5$), accuracy drops due to overly long contexts that hinder the selector. Overall, the framework is not highly sensitive to $K$, and since augmentation time grows roughly linearly while gains quickly saturate, a moderate choice such as $K=3$ offers the best trade-off between efficiency and effectiveness.

\stitle{Efficiency.}
Figure~\ref{fig:hyperparameters} also shows that the end-to-end construction of the augmented training set is computationally affordable: on a single-node server equipped with L20 GPUs and using vLLM for inference, the pipeline requires roughly 1–2 seconds per example on average. This overhead is modest, making the proposed trace-construction procedure feasible for deployment in real-world production settings.

\begin{table}[tbp]
    \scriptsize
    \centering
    \renewcommand{\arraystretch}{1.0} 
    \setlength{\tabcolsep}{2pt} 
    \caption{Cross-domain generalization.}\label{table:cross data} 
    \begin{tabular}{l|c|c|c|c}
    \toprule
     & \multicolumn{2}{c|}{In-domain generalization} & \multicolumn{2}{c}{Cross-domain generalization }\\
     \midrule
    Source (Tuning) & OK-VQA & FVQA  & OK-VQA & FVQA  \\
    \midrule 
    Target (Testing) & OK-VQA & FVQA & FVQA & OK-VQA \\
    \midrule
    Frozen$_{Qwen}$  &75.74 &71.61 &71.61 &75.74 \\
    SFT$_{Qwen}$     &76.36 \textcolor{mygreen}{(+0.62)} &73.91 \textcolor{mygreen}{(+2.30)} &64.77 \textcolor{myred}{(-6.84)} &67.50 \textcolor{myred}{(-8.24)} \\
    \model$_{Qwen}$ &91.51 \textcolor{mygreen}{(+15.77)}&\textbf{82.82} \textcolor{mygreen}{(+11.21)}&82.09 \textcolor{mygreen}{(+10.48)} &\textbf{85.45} \textcolor{mygreen}{(+9.71)} \\
    \hline
    Frozen$_{Llama}$  &67.84 &66.09 &66.09 &67.84\\
    SFT$_{Llama}$     &75.30 \textcolor{mygreen}{(+7.46)} &74.68 \textcolor{mygreen}{(+8.59)}&63.45 \textcolor{myred}{(-2.64)} &64.19 \textcolor{myred}{(-3.65)}\\
    \model$_{Llama}$ &90.01 \textcolor{mygreen}{(+22.17)} &\textbf{80.19} \textcolor{mygreen}{(+14.10)}&80.09 \textcolor{mygreen}{(+14.00)} &\textbf{79.59} \textcolor{mygreen}{(+11.75)} \\
    \hline
    Frozen$_{Gemma}$  &71.40 &70.64 &70.64 &71.40\\
    SFT$_{Gemma}$     &74.45 \textcolor{mygreen}{(+3.05)} &73.73 \textcolor{mygreen}{(+3.09)}&66.83 \textcolor{myred}{(-3.81)} &63.91 \textcolor{myred}{(-7.49)}\\
    \model$_{Gemma}$ &91.90 \textcolor{mygreen}{(+20.50)}&\textbf{81.20} \textcolor{mygreen}{(+10.56)} &81.20 \textcolor{mygreen}{(+10.56)} &\textbf{83.43} \textcolor{mygreen}{(+12.03)} \\
    \bottomrule
\end{tabular}
\end{table}

\subsection{Cross-domain Generalization}
Robust generalization to out-of-distribution (OOD) data is crucial in real-world applications.
We therefore evaluate both \emph{in-domain} and \emph{cross-domain} generalization across OK-VQA and FVQA, using three model variants: \textbf{Frozen} (backbone without fine-tuning), \textbf{SFT} (standard supervised fine-tuning), and our proposed \model, each instantiated with three MLLM backbones.
\textbf{In-domain generalization.}
When training and testing on the same dataset (OK-VQA or FVQA), both SFT and \model\ yield substantial gains over the Frozen model (left half of Table~\ref{table:cross data}), confirming that fine-tuning effectively adapts to the target domain.
Our framework further improves performance by explicitly supervising intermediate reasoning.
\textbf{Cross-domain generalization.}
We then examine transfer across datasets in both directions, OK-VQA $\rightarrow$ FVQA and FVQA $\rightarrow$ OK-VQA, which induces a substantial distribution shift.
As shown in the right half of Table~\ref{table:cross data}, SFT often degrades sharply—and in some cases even underperforms the Frozen baseline—highlighting vulnerability to catastrophic forgetting and limited cross-domain robustness.
In contrast, \model\ consistently avoids such degradation and even improves performance on the unseen domain, indicating stronger resistance to forgetting and superior cross-domain generalization.

\subsection{Qualitative Case Study}
As shown in \Cref{tab:case_example_1,tab:case_example_2}, \model\ differs from strong MLLMs by emitting \emph{trace-first} dual paths that expose key entity$\rightarrow$relation hops before answering.
In the OK-VQA example, Gemma-3-12B and Gemini~2.5~Pro both produce confident free-form narratives and predict the wrong ``Persian Gulf'', whereas \model\ follows a compact scaffold (\texttt{ship.hull\_number → ship.name → location.island\_group → ocean.name}) and outputs the annotator-favored \textbf{atlantic}.
In the FVQA example, baselines over-focus on the patterned ``rug'', while \model\ reasons via human–object \emph{affordances} and correctly selects \textbf{sofa}.
Overall, these suggest that \model\ reduces overconfident, opaque failure modes and texture-based shortcuts by grounding predictions in short, relation-based scaffolds linking visual cues and parametric knowledge to the final answer.

\begin{table}[tbp]
\scriptsize
\centering
\caption{Qualitative case study example, in OK\-VQA}
\label{tab:case_example_1}
\renewcommand{\arraystretch}{1.0}
\begin{tabular}{|>{\centering\arraybackslash}m{0.55\linewidth}|
                  >{\centering\arraybackslash}m{0.38\linewidth}|}
\hline
\makebox[\linewidth][c]{\includegraphics[width=\linewidth]{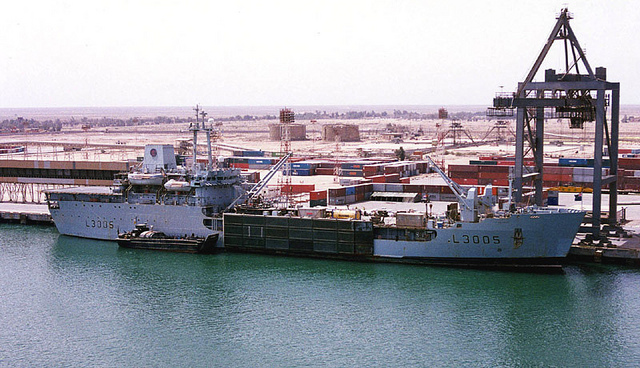}} &
\begin{minipage}[l]{\linewidth}                  
  \textbf{Question:} Name the sea shown in this picture where the ship is seen? \\[0.5mm]
  \textbf{Answers:} [atlantic, atlantic, atlantic, atlantic, dock of red sea, dock of red sea, kamchatka, kamchatka, philippine, philippine]
\end{minipage} \\
\hline


\rowcolor{tableheaderbg}
\multicolumn{2}{|l|}{\textbf{Answer from Gemma-3-12B}} \\
\multicolumn{2}{|p{1\linewidth}|}{
  \begin{minipage}[t]{1\linewidth}
    Here's a breakdown of the image and the answer to your question: 1. Object Identification and Attributes: * Ships: Two large, grey ships are prominently featured. They appear to be military or transport vessels, with the designation ``L3005" visible on their hulls. * Containers: Numerous shipping containers in various colors (red, green, grey) are stacked near the ships and along the docks. * Dock/Port Area: A large port or dock area is visible, with cranes, storage areas, and other infrastructure. * Crane: A large gantry crane is present, likely used for loading and unloading cargo. * Water: A body of water (sea or harbor) surrounds the ships and docks. * Landscape: A flat, arid landscape is visible in the background. * Weather: The sky is hazy and the lighting suggests a bright, sunny day. 2. Relationship Path \& Reasoning: The question asks ``Name the sea shown in this picture where the ship is seen?". To answer this, we need to consider the context of the image. The presence of a large port, military-style ships, and the arid landscape provide clues. Based on the visual cues, this is likely the Persian Gulf. The arid landscape, the presence of a large port facility, and the type of ships are all consistent with the Persian Gulf region. 3. Answer: The sea shown in this picture is the Persian Gulf.
  \end{minipage}
} \\
\hline

\rowcolor{tableheaderbg}
\multicolumn{2}{|l|}{\textbf{Answer from Gemini 2.5 Pro}} \\
\multicolumn{2}{|p{1\linewidth}|}{
  \begin{minipage}[t]{1\linewidth}
    Based on the visual evidence, the ship in the image can be identified, which helps pinpoint the location. The ship with the pennant number L3005 is the British Royal Fleet Auxiliary landing ship RFA Sir Galahad. This photograph was taken at the port of Umm Qasr, Iraq, likely around 2003 during humanitarian relief operations following the coalition invasion. The port of Umm Qasr is situated on the Khawr Abd Allah waterway, which flows into the Persian Gulf. Therefore, the sea connected to the port shown in the picture is the Persian Gulf.
  \end{minipage}
} \\
\hline

\rowcolor{tableheaderbg}
\multicolumn{2}{|l|}{\textbf{Answer from \model}} \\
\multicolumn{2}{|p{1\linewidth}|}{
  \begin{minipage}[t]{1\linewidth}
    \textbf{vision path:} ship.hull\_number $\rightarrow$ ship.name $\rightarrow$ location.island\_group\\
    \textbf{text path:} location.island\_group $\rightarrow$ location.ocean $\rightarrow$ ocean.name \\[0.0mm]
Here is one way to reason about this: 1. Identify the ship: The ship in the image has the hull number \texttt{L3005}, which identifies it as the Royal Fleet Auxiliary ship \emph{Sir Galahad}. 2. Link the ship to a location: RFA Sir Galahad is known for its role in the Falklands War and was sunk near the Falkland Islands. Thus we can link the scene to the island group ``Falkland Islands''. 3. Map the island group to a sea: The Falkland Islands are located in the South Atlantic Ocean. Therefore, the body of water in which this ship is seen is the South Atlantic (part of the Atlantic Ocean). 
\\[0.0mm]
Therefore, the possible answers include: \textbf{\textcolor{red}{atlantic}}, pacific
  \end{minipage}
} \\
\hline

\end{tabular}
\end{table}

\begin{table}[tbp]
\scriptsize
\centering
\caption{Qualitative case study example, in FVQA}
\label{tab:case_example_2}
\renewcommand{\arraystretch}{1.0}

\begin{tabular}{|>{\centering\arraybackslash}m{0.55\linewidth}|
                  >{\centering\arraybackslash}m{0.35\linewidth}|}
\hline

\makebox[\linewidth][c]{\includegraphics[width=\linewidth]{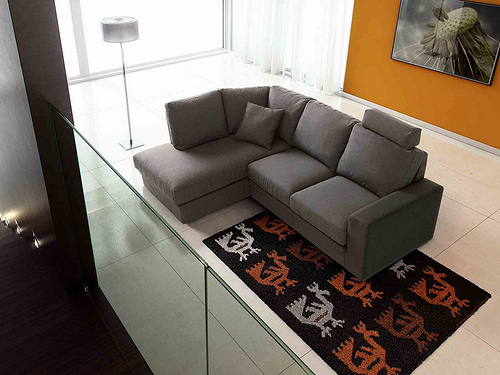}} &
\begin{minipage}[l]{\linewidth}                   
  \textbf{Question:} Which object in this image is used for play? \\[0.5mm]
  \textbf{Answer:} sofa
\end{minipage} \\
\hline

\rowcolor{tableheaderbg}
\multicolumn{2}{|l|}{\textbf{Answer from Gemma-3-12B}} \\
\multicolumn{2}{|p{1\linewidth}|}{%
  \begin{minipage}[t]{1\linewidth}
Based on the image, the object most likely used for play is the rug. Here's why: * Location: It's placed on the floor in a living room setting, a common area for play. * Pattern: The rug has a playful, abstract pattern that could be engaging for children. * Function: Rugs are often used as play surfaces for kids. Let me know if you'd like me to analyze the image for other details!
  \end{minipage}
} \\
\hline

\rowcolor{tableheaderbg}
\multicolumn{2}{|l|}{\textbf{Answer from Gemini 2.5 Pro}} \\
\multicolumn{2}{|p{1\linewidth}|}{%
  \begin{minipage}[t]{1\linewidth}
    Based on the image provided, there are no explicit toys or games visible. However, the object most likely to be used for play, especially imaginative play for a child, is the rug. The rug features a pattern of stylized, repeating figures (which look like birds or other animals). Children often use such patterns on floors or rugs as a landscape for their games, moving toys around them or creating stories about the figures.
  \end{minipage}
} \\
\hline

\rowcolor{tableheaderbg}
\multicolumn{2}{|l|}{\textbf{Answer from \model}} \\
\multicolumn{2}{|p{1\linewidth}|}{%
  \begin{minipage}[t]{1\linewidth}
    \textbf{vision path:} scene.room\_type $\rightarrow$ object.sofa $\rightarrow$ object.affordance.play \\
    \textbf{text path:} object.category $\rightarrow$ object.typical\_use $\rightarrow$ play.activity \\[0.5mm]
Based on the image, the most plausible object used for play is the sofa. Visually, the scene appears to be a living room with a large, soft sofa that offers comfortable space for sitting, lying down, or jumping around, which are typical play activities, especially for children. Other visible objects are less likely to be directly associated with play: the rug mainly serves as a floor covering, the floor lamp provides lighting, the artwork is decorative, and the glass railing is a structural element for safety rather than play. The sofa, by contrast, affords relaxed and playful behavior in this setting. \\[0.5mm]
Therefore, the possible answers include: \textbf{\textcolor{red}{sofa}}.
  \end{minipage}
} \\
\hline

\end{tabular}
\end{table}

%% file: sec/sec-appendix.tex
\section{Baselines}\label{app:baselines}
\stitle{KVQA with Knowledge Graphs and Retrieval.}
We select representative state-of-the-art approaches, including direct question-only answering (Q Only) \citep{Marino2019OKVQAAV}, BAN \citep{Kim2018BilinearAN}, MUTAN \citep{Benyounes2017MUTANMT}, ConceptBERT \citep{garderes2020conceptbert}, KRISP \citep{marino2021krisp}, MAVEx \citep{wu2022multi}, VLCBERT \citep{Ravi2022VLCBERTVQ}, HCNMN \citep{Zhang2023TowardMD}, and MCAN \citep{Yu2019DeepMC}. Since BAN and MUTAN are limited to learning unimodal visual features, we enhance them with ArticleNet (AN) \citep{Marino2019OKVQAAV}, which retrieves relevant information from Wikipedia based on the given question–image pair to support external knowledge reasoning. These enhanced versions are referred to as “BAN + AN” and “MUTAN + AN” \citep{Marino2019OKVQAAV}.

\stitle{KVQA with LLMs \ / MLLMs.}
We employ PICa \citep{Yang2021AnES}, KAT \citep{Gui2021KATAK}, and REVIVE \citep{Lin2022REVIVERV}.
The results of \textit{KVQA with Knowledge Graphs and Retrieval} as well as \textit{KVQA with Large Language Models} are from prior work \citep{dong-etal-2024-modality}, where the exact same experimental setup and evaluation protocols are adopted.

\stitle{IK-KVQA with Multimodal Large Language Models.} 
We employed three types of Multimodal Large Language Models (MLLMs):
\begin{itemize}
    \item \textbf{Advanced open-source MLLMs}: Including three regular-sized models: Qwen2.5-VL-7B \citep{Bai2025Qwen25VLTR}, Llama-3.2-11B-Vision \citep{Dubey2024TheL3}, and Gemma-3-12B \citep{Kamath2025Gemma3T}; as well as three larger and more advanced models: Gemma-3-27B \citep{Kamath2025Gemma3T}, Qwen2.5-VL-72B \citep{Bai2025Qwen25VLTR}, and InternVL3-78B \citep{Zhu2025InternVL3EA}. All of them are \textbf{instruction-tuned versions}.
    \item \textbf{Proprietary state-of-the-art MLLMs}: Including two of Google’s most advanced models, Gemini 2.5 Flash and Gemini 2.5 Pro \citep{Comanici2025Gemini2P}, as well as OpenAI’s flagship multimodal model, GPT-4o \citep{Hurst2024GPT4oSC}. Both Gemini 2.5 Flash and Gemini 2.5 Pro perform inference in the Dynamic Thinking mode.
    \item \textbf{Augmented MLLMs}: 
    \begin{itemize}
        \item \textbf{Supervised fine-tuning} (\textbf{SFT}) \citep{ouyang2022training}  is a crucial process that trains a pre-trained MLLM on a high-quality dataset of instructions and responses, making it more effective at following specific commands and performing user-facing tasks. The MLLM backbone is Qwen2.5-VL-7B.
        \item \textbf{Chain of Thought (CoT)} \citep{wei2022chain} is a prompting technique that improves the reasoning abilities of large language models by guiding them to break down a complex problem into a series of intermediate steps before providing a final answer. The MLLM backbone is Qwen2.5-VL-7B.
        \item \textbf{CoT + SFT} is a well-optimized CoT-prompted SFT baseline. 
        \item  \textbf{LLaVA-CoT} \citep{xu2024llava}, a new multimodal model that uses a chain-of-thought method to improve vision-language models' ability to reason step-by-step.
        \item \textbf{M2-Reasoning} (7B) \citep{ai2025m2} is a multimodal large language model (MLLM) that achieves state-of-the-art (SOTA) performance in both general and spatial reasoning by using a high-quality data pipeline and a dynamic multi-task training strategy.
        \item \textbf{ Self-Distillation Fine-Tuning (SDFT)} \citep{Yang2024SelfDistillationBD} rewrites task responses into its own style and fine-tunes on them to reduce distribution shift and forgetting. The MLLM backbone is Qwen2.5-VL-7B.
    \end{itemize}
\end{itemize}

\subsection{Implementation Details.}
\label{app:baseline:details}
Our approach \model\ has been implemented using PyTorch
2.7.0 as well as Python 3.10, and all experiments have been conducted on the NVIDIA L20 GPU.
During training, the batch size (with accumulation) is set to $16$, the learning rate is $1\mathrm{e}{-4}$, the LoRA rank is 32, the LoRA alpha is 64, the traininig epoch is 3.
In the OK-VQA dataset, $K$ is set as 3, and in the FVQA dataset, $K$ is set as 4.

We adhere to the established evaluation setting and fix the random seed to \texttt{42} throughout data loading, parameter initialization, and decoding.
Consistent with prior work \citep{dong-etal-2024-modality}, we report single-run results in the main tables to maintain strict comparability with published baselines.
We did not sweep over seeds or report standard deviations; we view multi-seed evaluation as complementary and leave it to future extensions or large-scale replication studies.

To ensure a level playing field across closed- and open-source models, we (i) supply only the image and the question as inputs, without chain-of-thought or auxiliary prompts; and (ii) adopt each model's \emph{default} inference hyperparameters (decoding temperature and maximum generation length), avoiding any model-specific tuning.
This protocol matches the default settings recommended by the model providers and prevents gains from hyperparameter overfitting.

\section{Metric} \label{app: metric}
For the open-ended task, \ie, direct answer (DA) setting, we evaluate generated answers using the following accuracy definition:
\begin{align}
    \text{Accuracy} = \min \left( \frac{\# \text{humans that provided that answer}}{3}, \; 1 \right)
\end{align}

\noindent
\textit{i.e.}, an answer is considered fully correct (100\% accuracy) if it matches the responses of at least three annotators. Before comparison, all responses are normalized by lowercase, converting numbers to digits, and removing punctuation and articles. We deliberately avoid soft similarity measures such as Word2Vec~\citep{Mikolov2013DistributedRO}, which may incorrectly cluster semantically distinct words (e.g., ``left'' vs. ``right''). Likewise, we exclude machine translation metrics such as BLEU and ROUGE, as they are mainly suited for multi-word sentence evaluation rather than short answers typically found in VQA.

\section{Theoretical Notes for StaR-KVQA}
\label{app:theory}

This appendix offers compact analyses that formalize how (i) typed, path-grounded traces (planner + \emph{reasoning composer}), (ii) the single-model selector, and (iii) single-model self-distillation contribute to StaR-KVQA. The statements are backbone-agnostic and match the components introduced in Sec 4.

\subsection{Notation and Standing Assumptions}
Let $(I,Q,a^\star)\sim \mathcal{D}$ denote image, question, and ground-truth answer. 
A \emph{trace} is $T=(P_t,P_v,C)$. 
Our model with parameters $\theta$ induces
\begin{align}
p_\theta(T,a\,|\,I,Q)
&= p_\theta(P_t,P_v\,|\,I,Q)\;p_\theta(C\,|\,I,Q,P_t,P_v) \nonumber\\
&\quad p_\theta(a\,|\,I,Q,T).
\end{align}
We reuse two structural predicates from Sec.~4.2:
\begin{align}
\mathrm{Cover}(C;P_t,P_v)\ge \kappa,\qquad  \mathrm{Vis}(C;I)\ge \rho,
\end{align}
encoding path–sentence coverage and visual attestability. 
Define the feasible set $\mathcal{T}_{\kappa,\rho}=\{T:\mathrm{Cover}\ge \kappa,\ \mathrm{Vis}\ge \rho\}$.

\subsection{Generalization Benefit from Typed and Verifiable Traces}
\label{app:gen}
We compare an \emph{answer-only} class with a \emph{trace-constrained} class that must produce $T\in\mathcal{T}_{\kappa,\rho}$ alongside $a$.

\paragraph{Hypothesis classes.}
Let $\mathcal{H}_{\mathrm{ans}}=\{h:(I,Q)\mapsto a\}$ and
\begin{align}
\mathcal{H}_{\mathrm{trace}}=\{h:(I,Q)\mapsto (T,a)\ \text{s.t.}\ T\in\mathcal{T}_{\kappa,\rho}\}.
\end{align}
Both are realized by the \emph{same} architecture but trained with different supervision.

\begin{theorem}[Rademacher shrinkage via verifiable structure]
\label{thm:rad}
Assume bounded losses $\ell(a,a^\star)\in[0,1]$ and $\ell_{\mathrm{trace}}(T,a;a^\star)\in[0,1]$ with
$\ell_{\mathrm{trace}}(T,a;a^\star)\ge \ell(a,a^\star)$ and equality whenever $T\in\mathcal{T}_{\kappa,\rho}$. 
Then for any sample size $N$ and $\delta\in(0,1)$, with probability at least $1-\delta$,
\begin{align}
\mathcal{R}_{\mathcal{D}}(h_{\mathrm{trace}})
\le
\widehat{\mathcal{R}}_N(h_{\mathrm{trace}})+2\,\mathfrak{R}_N(\mathcal{H}_{\mathrm{trace}})
+\sqrt{\tfrac{\ln(1/\delta)}{2N}},
\end{align}
and moreover 
$\mathfrak{R}_N(\mathcal{H}_{\mathrm{trace}})\le \mathfrak{R}_N(\mathcal{H}_{\mathrm{ans}})\cdot \sqrt{\Pi(\mathcal{T}_{\kappa,\rho})/\Pi(\mathcal{T})}$,
where $\mathfrak{R}_N(\cdot)$ is the empirical Rademacher complexity and $\Pi(\cdot)$ the growth function.
\end{theorem}

\noindent\textit{Intuition.} Enforcing typed, verifiable traces prunes implausible labelings (fewer admissible traces per example), which lowers the effective complexity term and tightens the bound.  
\textit{Practical takeaway.} Structure acts as an inductive bias without changing the backbone.

\subsection{Selector as Maximum Likelihood under a Consistency-Noise Model}
\label{app:selector}
Our best-triplet selector uses the \emph{single-model} setup to score candidates. The score can be interpreted as a log-likelihood under a simple noise model.

\paragraph{Model.}
For candidate $b$, define binary indicators
$Y_b^{(\text{ans})},Y_b^{(\text{ent})},Y_b^{(\text{align})},Y_b^{(\text{coh})}\in\{0,1\}$ 
for answer correctness, explanation$\Rightarrow$answer entailment, path$\rightarrow$explanation alignment, and explanation coherence. 
Assume conditional independence given a latent quality $q_b$:
\begin{align}
\Pr(Y_b^{(j)}=1\mid q_b)=\sigma(w_j q_b),\qquad j\in\{\text{ans, ent, align, coh}\},
\end{align}
with logistic $\sigma$ and weights $w_j>0$. 
Let $\hat{y}_b^{(j)}\in[0,1]$ be soft proxies estimated by the model; the log-likelihood is
$\log L_b(q_b)=\sum_j \hat{y}_b^{(j)}\log \sigma(w_j q_b) + (1-\hat{y}_b^{(j)})\log (1-\sigma(w_j q_b))$.

\begin{proposition}[Selector equals MLE/MAP ranking]
\label{prop:mle}
The maximizer $\hat q_b=\arg\max_{q} \log L_b(q)$ is monotone in 
$s_\phi(b):=\sum_j w_j (2\hat{y}_b^{(j)}-1)$.
Therefore selecting $b^\star=\arg\max_b s_\phi(b)$ agrees with MLE (and with MAP under any log-concave prior).
\end{proposition}

\noindent\textit{Intuition.} The weighted consistency cues act like independent “votes.” A larger weighted sum implies a larger MLE quality and thus a higher rank.  
\textit{Practical takeaway.} Our LLM-as-a-judge ranking matches likelihood-based selection under a reasonable noise model.

\subsection{Single-Model Self-Distillation Reduces Supervision–Generation Shift}
\label{app:sd}
Let $P$ be the generator distribution over traces (from $\mathrm{MLLM}_\phi$) and $Q_\theta$ the student’s distribution after fine-tuning. 
Let $\mathcal{L}\in[0,1]$ be a bounded loss on completions.

\begin{lemma}[Risk gap upper bounded by divergence]
\label{lem:pinsker}
For any $(I,Q)$,
\begin{align}
\big|\,\mathbb{E}_{T\sim P}\mathcal{L}(T)-\mathbb{E}_{T\sim Q_\theta}\mathcal{L}(T)\,\big|
\;\le\; \sqrt{2\,\mathrm{KL}\!\left(P\,\|\,Q_\theta\right)}.
\end{align}
\end{lemma}

\begin{proof}
By total variation (TV) and Pinsker’s inequality: 
$|\,\mathbb{E}_P f-\mathbb{E}_Q f\,|\le 2\,\mathrm{TV}(P,Q)$ for $f\in[0,1]$, and 
$\mathrm{TV}(P,Q)\le \sqrt{\tfrac12 \mathrm{KL}(P\|Q)}$. 
Combining gives the stated bound.
\end{proof}

\begin{theorem}[Self-distillation alignment]
\label{thm:sd}
If fine-tuning reduces $\mathrm{KL}(P\|Q_\theta)$ on augmented traces (i.e., the student learns from traces in the generator’s style), the supervision–generation risk gap is $O(\sqrt{\mathrm{KL}(P\|Q_\theta)})$ by Lemma~\ref{lem:pinsker}. 
Using a \emph{single-model} setup (shared format/tokenization) typically attains a smaller KL than heterogeneous teachers.
\end{theorem}

\noindent\textit{Intuition.} Learning from “in-style” traces narrows the distribution gap, which directly controls the risk gap.  
\textit{Practical takeaway.} Single-model self-distillation stabilizes training and mitigates forgetting.

\subsection{Training Objective as a Joint-Likelihood Lower Bound}
\label{app:elbo}
Our loss in Sec.~4 supervises $(P_t,P_v)$, $C$, and $a$. It can be seen as maximizing a lower bound on $\log p_\theta(a^\star\,|\,I,Q)$ marginalized over feasible traces.

\begin{proposition}[ELBO-style lower bound with feasible traces]
\label{prop:elbo}
Let $\mathcal{T}_{\kappa,\rho}$ be the feasible set. 
For any auxiliary distribution $q(T\,|\,I,Q)$ supported on $\mathcal{T}_{\kappa,\rho}$,
\begin{equation}
\begin{aligned}
\log p_\theta(a^\star\,|\,I,Q)
\;\ge\;
&\underbrace{\mathbb{E}_{q}\!\left[\log p_\theta(P_t,P_v\,|\,I,Q)\right]}_{\text{path term}}\\
&+\underbrace{\mathbb{E}_{q}\!\left[\log p_\theta(C\,|\,I,Q,P_t,P_v)\right]}_{\text{explanation term}}\\
&+\underbrace{\mathbb{E}_{q}\!\left[\log p_\theta(a^\star\,|\,I,Q,T)\right]}_{\text{answer term}}\\
&-\mathrm{KL}\!\left(q(T\,|\,I,Q)\,\|\,p_\theta(T\,|\,I,Q,a^\star)\right).
\end{aligned}
\end{equation}

\end{proposition}

\begin{proof}
Write $\log p_\theta(a^\star\,|\,I,Q)=\log \sum_{T\in \mathcal{T}_{\kappa,\rho}} p_\theta(T,a^\star\,|\,I,Q)$, insert $q(T\,|\,I,Q)$, and apply Jensen:
\[
\log \sum_T q(T)\frac{p_\theta(T,a^\star)}{q(T)}
\ge \mathbb{E}_{q}\big[\log p_\theta(T,a^\star)-\log q(T)\big].
\]
Factorize $p_\theta(T,a^\star)$ using the model and rearrange.
\end{proof}

\noindent\textit{Intuition.} Supervising paths, explanations, and answers maximizes a tractable surrogate of the marginal likelihood; better selection of $q$ (stronger traces) tightens the bound.  
\textit{Practical takeaway.} Improving the selector/feasibility checks translates into better training signals.

\subsection{Putting Pieces Together}
Theorems~\ref{thm:rad}–\ref{thm:sd} and Prop.~\ref{prop:elbo} jointly suggest: 
(i) typed, verifiable traces reduce effective hypothesis space; 
(ii) the single-model selector is equivalent to MLE/MAP under a simple consistency–noise view; 
(iii) single-model self-distillation reduces supervision–generation shift; and 
(iv) the training objective maximizes a joint-likelihood lower bound whose tightness benefits from stronger traces and selection.

\section{Use of Large Language Models}
In preparing this article, Large Language Models (LLMs) were employed only for stylistic refinement. Their role was limited to editing the wording of certain sections in order to improve readability and fluency of the manuscript. The intellectual contributions—including the development of ideas, design of experiments, analysis of results, and formulation of conclusions—were carried out entirely by the authors.
No part of the research process, data interpretation, or scientific claims relied on the use of LLMs. The authors assume full responsibility for the content presented and ensure its originality and accuracy.

\section{Data Ethics Statement}
To evaluate the efficacy of \model, we conducted experiments which only use publicly available
datasets, namely, OK-VQA \citep{Marino2019OKVQAAV} and FVQA \citep{Wang2016FVQAFV}.
We also confirm that no personally identifiable information was utilized, and this research did not involve any human or animal subjects.



